\documentclass[10pt,letterpaper]{article}

\usepackage{cogsci}

\cogscifinalcopy 

\usepackage{pslatex}
\usepackage{apacite}
\usepackage{float} 

\usepackage{subcaption}
\usepackage{graphicx}
\usepackage{amsmath}
\usepackage{amsfonts}
\usepackage{booktabs} 
\usepackage{multirow}

\usepackage[switch]{lineno}

\usepackage{comment}

\usepackage{soul} 
\usepackage{color}
\usepackage{xcolor,colortbl}
\definecolor{orange}{HTML}{FCBF64}
\definecolor{skin}{HTML}{E79E97}
\definecolor{lightblue}{HTML}{ADD8E6}

\title{Modeling Question Asking Using Neural Program Generation}
\author{{\large \bf Ziyun Wang (ziyunw@nyu.edu)} \\
  Department of Computer Science \\
  New York University\thanks{Ziyun is now at Tencent.}
  \And {\large \bf Brenden M. Lake (brenden@nyu.edu)} \\
  Department of Psychology and Center for Data Science \\
  New York University}

\begin{document}
\maketitle

\begin{abstract}
People ask questions that are far richer, more informative, and more creative than current AI systems. We propose a neuro-symbolic framework for modeling human question asking, which represents questions as formal programs and generates programs with an encoder-decoder based deep neural network. From extensive experiments using an information-search game, we show that our method can predict which questions humans are likely to ask in unconstrained settings. We also propose a novel grammar-based question generation framework trained with reinforcement learning, which is able to generate creative questions without supervised human data.

\end{abstract}

\section{Introduction}

People can ask rich, creative questions to learn efficiently about their environment. Question asking is central to human learning yet it is a tremendous challenge for computational models. There is an infinite set of possible questions that one can ask, leading to challenges both in representing the space of questions and in searching for the right question to ask.

Machine learning has been used to address aspects of this challenge.
Traditional methods have used heuristic rules designed by humans \cite{heilman2010good,chali2015towards}, which are usually restricted to a specific domain. Recently, neural network approaches have also been proposed, including retrieval methods which select the best question from past experience \cite{Mostafazadeh2016GeneratingNQ} and encoder-decoder frameworks which map visual or linguistic inputs to questions \cite{serban2016generating,yao2018teaching}.
While effective in some settings, these approaches do not consider settings where the questions are asked about partially unobservable states. Furthermore, these methods are heavily data-driven, limiting the diversity of generated questions and requiring large training sets for different goals and contexts. There is still a large gap between how people and machines ask questions.

\begin{figure}[t]
    \centering
    \includegraphics[width=0.49\textwidth]{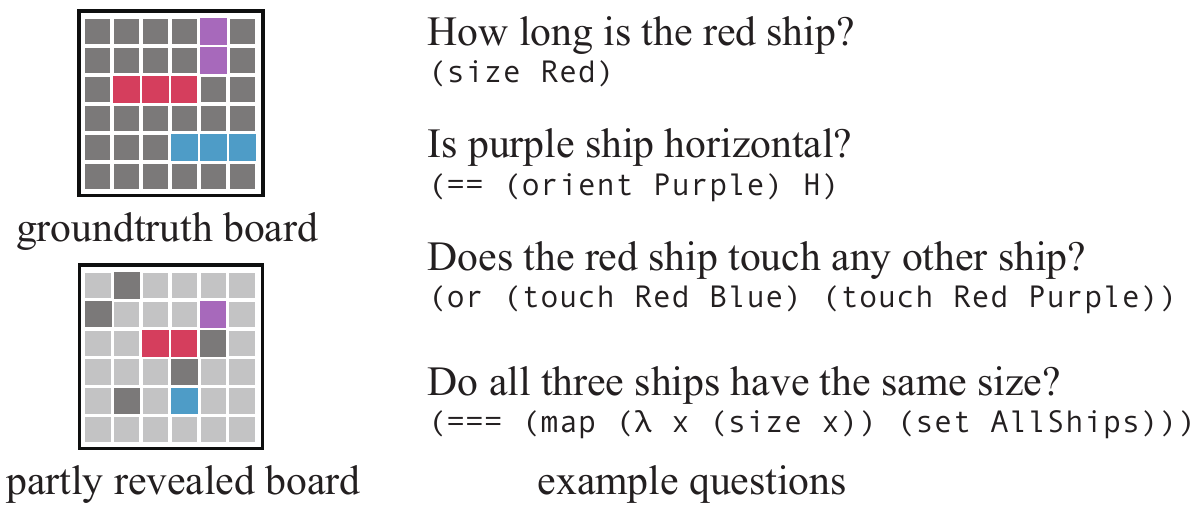}
    \caption{The Battleship task. Blue, red, and purple tiles are ships, dark gray tiles are water, and light gray tiles are hidden. The agent can see a partly revealed board, and should ask a question to seek information about the hidden board. Example questions and translated programs are shown on the right. We recommend viewing the figures in color.}
    \label{fig:task}
\end{figure}

Recent work has aimed to narrow this gap by taking inspiration from cognitive science. For instance, \citeA{Lee2018Answerer} incorporates aspects of ``theory of mind'' \cite{premack1978does} in question asking by simulating potential answers to the questions. \citeA{rao2019answer} propose a similar method which estimates the value of an answer using neural networks. But these approaches rely on imperfect natural language understanding which may lead to error propagation. Related to our approach, \citeA{rothe2017question} proposed a question-asking framework by modeling questions as symbolic programs, but their algorithm relies on hand-designed program features and requires expensive calculations.

We use ``neural program generation'' to bridge symbolic program generation and deep neural networks, bringing together some of the best qualities of both approaches. Symbolic programs provide a compositional ``language of thought'' \cite{fodor1975lot} for creatively synthesizing which questions to ask, allowing the model to construct new ideas based on familiar building blocks. Compared to natural language, programs are precise in their semantics, have clearer internal structure, and require a much smaller vocabulary, making them an attractive representation for question answering systems as well \cite{johnson2017inferring,mao2019neuro}.
However, there has been much less work using program synthesis for question asking, which requires searching through infinitely many questions (where many questions may be informative) rather than producing a single correct answer to a question. Deep neural networks allow for rapid question-synthesis using encoder-decoder modeling, eliminating the need for the expensive symbolic search and feature evaluations in \citeA{rothe2017question}. Together, the questions can be synthesized quickly and evaluated formally for quality (e.g. the expected information gain), which as we show can be used to train question asking systems using reinforcement learning.

In this paper, we develop a neural program generation model for asking questions in an information-search game, which is similar to ``Battleship'' and has been studied previously \cite{gureckis2009active,rothe2017question,Rothe2018}. Our model uses a convolutional encoder to represent the game state, and a Transformer decoder for generating questions in a domain specific language (DSL). 
Importantly, we show that the model can be trained from a small number of human demonstrations of good questions, after pre-training on a large set of automatically generated questions. Our model can also be trained without such demonstrations using reinforcement learning, while still expressing important characteristics of human behavior.
We evaluate the model on two aspects of human question asking, including density estimation based on free-form question asking, and creative generation of genuinely new questions. 

To summarize, our paper makes three main contributions: 1) We propose a neural network for modeling human question-asking behavior, 2) We propose a novel reinforcement learning framework for generating creative human-like questions by exploiting the power of programs, and 3) We evaluate different properties of our methods extensively through experiments.

\section{Related work}
Question generation has attracted attention from the machine learning community. Early research mostly explored rule-based methods which strongly depend on human-designed rules \cite{heilman2010good,chali2015towards}. Recent methods for question generation adopt deep neural networks, especially using the encoder-decoder framework, and can generate questions without hand-crafted rules. These methods are mostly data-driven, which use pattern recognition to map inputs to questions. Researchers have worked on generating questions from different types of inputs such as knowledge base facts \cite{serban2016generating}, pictures \cite{Mostafazadeh2016GeneratingNQ}, and text for reading comprehension \cite{yuan2017machine,yao2018teaching}.
However aspects of human question-asking remain beyond reach, including the goal-directed and flexible qualities that people demonstrate when asking new questions. This issue is partly addressed by recent papers that draw inspiration from cognitive science.
Research from \citeA{rothe2017question} and \citeA{Lee2018Answerer} generate questions by sampling from a candidate set based on goal-oriented metrics. This paper introduces an approach to question generation that does not require a candidate question set and expensive feature computations at inference time. Moreover, our approach can learn to ask good questions without human examples through reinforcement learning. 

Another task that shares some similarity with our setting is image captioning, where neural networks are employed to generate a natural language description of an image. Typical approaches use a CNN-based image encoder and a recurrent neural network (RNN) \cite{mao2014explain, vinyals2015show, karpathy2015deep} or a Transformer \cite{yu2019multimodal} as a decoder for generating caption text. The model design often resembles ours, with important differences. First, we use programs as output rather than text. Second, our question-asking setting requires deeper reasoning about entities in the input than is typically required in captioning datasets.

Our work also builds on neural network approaches to program synthesis, which have been applied to many different domains \cite{devlin2017robustfill,tian2019learning}.
Those approaches often draw inspiration from computer architecture, using neural networks to simulate stacks, memory, and controllers in differentiable form \cite{reed2015neural}. 
Other models incorporate Deep Reinforcement Learning (DRL) to optimize the generated programs in a goal oriented environment, such as generating SQL queries which can correctly perform a specific database processing task \cite{sen2020athena++}, translating strings in Microsoft Excel sheets \cite{devlin2017robustfill}, understanding and constructing 3D scenes \cite{liu2019learning} and objects \cite{tian2019learning}. 
Recent work has also proposed ways to incorporate explicit grammar information into the program synthesis process. \citeA{yin2017syntactic} design a module to capture the grammar information as a prior, which can be used during generation. Some recent papers \cite{Si2019LearningAM, alon2020structural} encode grammar with neural networks and use DRL to explicitly encourage the generation of semantically correct programs.
Our work differs from these in two aspects. First, our goal is to generate informative human-like questions in the new domain instead of simply correct programs. Second, we more deeply integrate grammar information in our framework, which directly generates programs based on the grammar.

\begin{figure*}[t]
    \centering
    \includegraphics[width=0.87\textwidth]{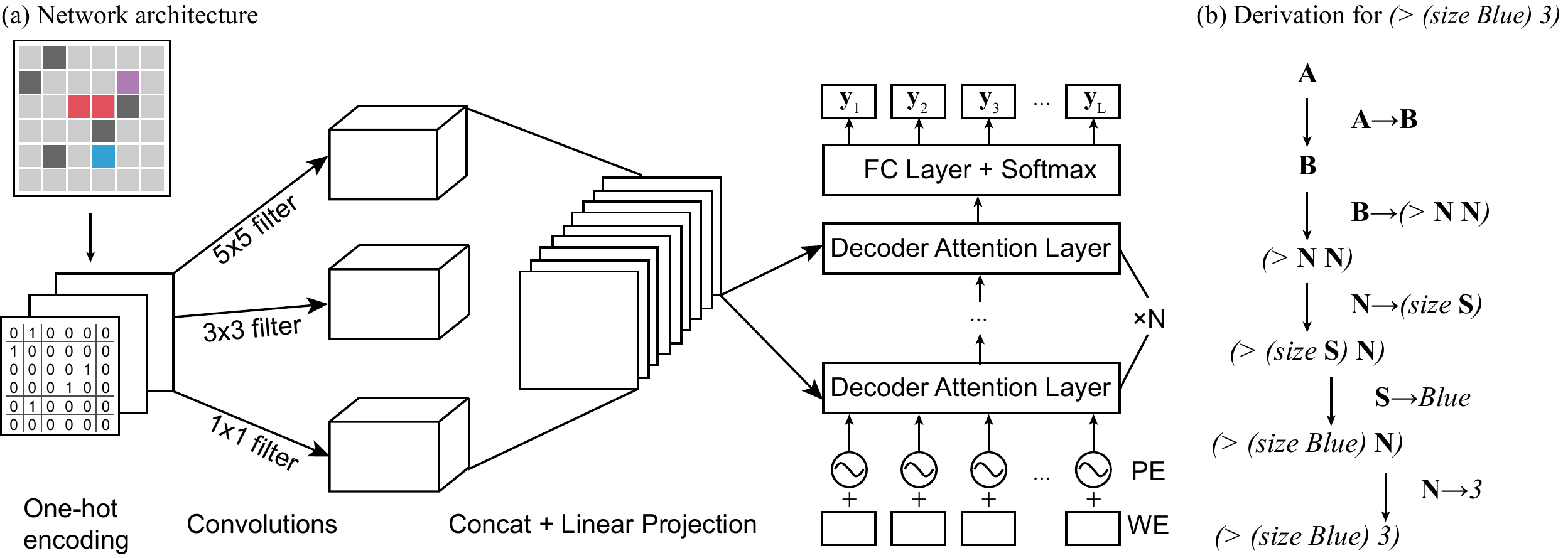}
    \caption{Neural program generation. Figure (a) shows the network architecture. The board is represented as a grid of one-hot vectors and is embedded with a convolutional neural network. The board embedding and a sequence of symbols are inputted to a Transformer decoder to generate output vectors. PE means positional embeddings, and WE means word embeddings. (b) shows the derivation steps for program ``{\tt (> (size Blue) 3)}'' using a context-free grammar. Non-terminals are shown as bold-faced, and terminals are shown in italic. The production rules used are shown next to each arrow.}
    \label{fig:full_model}
\end{figure*}

\section{Battleship Task}
In this paper, we work with a task used in previous work for studying human information search \cite{gureckis2009active} as well as question asking \cite{Rothe2018}. The task is based on an information search game called ``Battleship'', in which a player aims to resolve the hidden layout of the game board based on the revealed information (Figure \ref{fig:task}). There are three ships with different colors (blue, red, and purple) placed on a $6\times6$ grid. Each ship is either horizontal or vertical, and is either $2$, $3$ or $4$ tiles long. All tiles are initially turned over (light grey in Figure \ref{fig:task}), and the player can flip one tile at a time to reveal an underlying color (either a ship color, or dark grey for water). The goal of the player is to determine the configuration of the ships (positions, sizes, orientations) in the least number of flips.

In the modified version of this task studied in previous work \cite{rothe2017question,Rothe2018}, the player is presented with a partly revealed game board, and is required to ask a natural language question to gain information about the underlying configuration. As shown in Figure \ref{fig:task}, the player can only see the partly revealed board, and might ask questions such as ``How long is the red ship?'' In this paper, we present this task to our computational models, requesting they generate questions about the game board.

The dataset from \citeA{rothe2017question} consists of a small set of human questions along with their semantic parses, which represent the questions as LISP-like programs that can be executed on the game states to get answers to the questions. The DSL consists of primitives (like numbers, ship colors, etc.) and functions (like arithmetic operators, comparison operators, and other board-related functions) from which questions can be composed. Here, we aim to synthesize questions in this ``language of thought'' of semantic forms, since it captures key notions of compositionality and computability. Figure \ref{fig:task} shows some examples of produced programs.

\section{Neural program generation framework} \label{sec:method}
The neural network we use includes a Convolutional Neural Network (CNN) for encoding the input board, and a Transformer decoder for estimating the symbol distribution, or selecting actions during reinforcement learning.
Figure \ref{fig:full_model}(a) illustrates the architecture.

\paragraph{Network Architecture}\label{sec:architecture}
The neural network we use includes a Convolutional Neural Network (CNN) for encoding the input board, and a Transformer decoder for estimating the symbol distribution or selecting actions. The input $x\in\{0,1\}^{6\times6\times5}$ is a binary representation of the 6x6 game board with five channels, one for each color to be encoded as a one-hot vector in each grid location. A simple CNN maps the input $x$ to the encoder output $\mathbf{e}\in\mathbb{R}^{6\times6\times M}$, where $M$ is the length of encoded vectors. Then a Transformer decoder takes $\mathbf{e}$ and a sequence of length $L$ as input, and outputs a sequence of vectors $\mathbf{y}_i\in\mathbb{R}^{N_o},i=1\cdots L$, where $N_o$ is the output size. The model is shown in Figure \ref{fig:full_model}(a).

\paragraph{Training}
Our model is compatible with both supervised and reinforcement learning. In the supervised setting, the goal is to model the distribution of questions present in the training set. Each output $\mathbf{y}_i\in\mathbb{R}^{N_o}$ is a symbol at position $i$ in the program, where $N_o$ is the number of different symbols in the grammar. 
The model is trained with a symbol-level cross entropy loss, and can be used to calculate the log-likelihood of a given sequence, or to generate a question symbol-by-symbol from left to right.

Additionally, we propose a novel grammar-enhanced RL training procedure for the framework. Figure \ref{fig:full_model}(b) illustrates the process of generating a program from the context-free grammar that defines the DSL (full specification in \citeA{rothe2017question}). Beginning from the start symbol ``{\tt A}'', at each step a production rule is picked and applied to one of the non-terminals in the current string. The choice of rule is modeled as a Markov Decision Process, and we solve it with DRL. Each state is a partially derived string passed to the decoder, and we use the first output $\mathbf{y}_1\in\mathbb{R}^{N_o}$ to represent the probability of selecting each production rule from the allowed rules over all $N_o$ rules. After the rule is applied, the new string is passed back into the decoder, repeating until the string has only terminals.
We adopt the leftmost derivation to avoid ambiguity, so at each step the left-most non-terminal is replaced.

\section{Experiments}

\subsection{Estimating the distribution of human questions} \label{sec_exp_human_q}
In this experiment, we examine if neural program generation can capture the distribution of questions that humans ask, using a conditional language model. We seek to avoid training on a large corpus of human-generated questions; the dataset only offers about 600 human questions on a limited set of 18 board configurations \cite{rothe2017question}, and importantly people can ask intelligent questions in a novel domain with little direct training experience. We design a two-step training process; first we pre-train the model on automatically generated questions using an informativeness measure, second we fine-tune the model using only small set of real human questions collected by \citeA{rothe2017question}.

To create pre-training data, we generate a large number of game boards and sample $K$ questions for each board. To generate boards, we uniformly sample the configuration of three ships and cover up an arbitrary number of tiles, with the restriction that at least one ship tile is observed. With a tractable game model and domain-specific language of questions, we can sample $K$ programs for each board based on the expected information gain (EIG) metric, which quantifies the expected information received by asking the question. EIG is formally defined as the expected reduction in entropy, averaged over possible answers to a question $x$,
\begin{equation}
    \text{EIG}(x)=\mathbb{E}_{d\in A_x}[I(p(h))-I(p(h|d;x))]
\end{equation}
where $I(\cdot)$ is the Shannon entropy. The terms $p(h)$ and $p(h|d;x)$ are the prior and posterior distribution of a possible ship configuration $h$ given question $x$ and answer $d \in A_x$. We generate 2,000 boards, and sample 10 questions for each board, with the log-probability of a question proportional to its EIG value, up to a normalizing constant and with a maximum length of 80 tokens per program. After pre-training, the model is fine-tuned on the small corpus of human questions for 15 epochs with the same process and hyperparameters as in the pre-training step.

\begin{table}
\centering
\footnotesize
\caption{Predicting which questions people ask. LL\_all, LL\_highEnt, and LL\_lowEnt shows mean log-likelihood across held out boards. High/low refers to a board's entropy.}
\begin{tabular}{lrrr} \toprule
Model & LL\_all & LL\_highEnt & LL\_lowEnt \\ \midrule
\multicolumn{2}{l}{\emph{Rothe et al. (2017)}~~~-1400.06} & - & - \\
full model & \textbf{-150.38$\pm$0.51} & -150.92$\pm$1.51 & \textbf{-156.38$\pm$1.92} \\
~~-pretrain & -242.53$\pm$2.32 & -260.84$\pm$4.99 & -249.98$\pm$3.87 \\
~~-finetune & -415.32$\pm$0.95 & -443.03$\pm$1.33 & -409.01$\pm$1.21 \\
~~-encoder & -153.50$\pm$0.41 & \textbf{-149.69$\pm$0.54} & -163.13$\pm$0.82 \\  \bottomrule
\end{tabular}
\label{tab:exp2_result}
\end{table}

\begin{table*}
\centering
\small
\caption{Analysis of question generation. The models are compared in terms of average EIG value, the ratio of EIG value greater than 0.95 or 0, number of unique and novel questions generated (by ``novel'' we mean questions not present in the dataset of human-asked questions). The EIG for the text-based model is computed based on the program form of the questions. The average length is the number of words for text-based model, and number of tokens for others. SP means step penalty.}
\begin{tabular}{lrrrrrr} \toprule
Model & avg. EIG & EIG\textgreater{}0.95 & EIG\textgreater{}0 & \#unique & \#unique novel & avg. length \\ \midrule
text-based & 0.928 & 62.80\% & 76.95\% & - & - & 8.21 \\ 
supervised & 0.972 & 45.70\% & 81.80\% & 183 & 103 & 10.98 \\
grammar RL (noSP) & 1.760 & {88.30\%} & {92.40\%} & 224 & 213 & 9.97 \\
grammar RL (SP0.02) & \textbf{1.766} & \textbf{94.90\%} & \textbf{96.80\%} & 111 & 96 & 6.03 \\
grammar RL (SP0.05) & 1.559 & 91.90\% & 95.20\% & 57 & 47 & 5.79 \\
\bottomrule
\end{tabular}
\label{tab:exp3_result}
\end{table*}

\subsubsection{Results and discussion}
To evaluate the model, we follow the same procedure as \citeA{rothe2017question}, where they run leave-one-out cross-validation on 16 different boards\footnote{\citeA{rothe2017question} only evaluate on 16 out of all 18 boards because the other two boards are too computationally expensive for their method as they mentioned in the paper.} and calculate the sum of log-likelihood of all questions for a board and average across different boards. We evaluate the log-likelihood of reference questions on our full model as well as lesioned variants, including a model without pre-training, a model without fine-tuning, and a model with only a decoder  (unconditional language model). A summary of the results is shown in Table \ref{tab:exp2_result}. We run each experiment $10$ times and report the mean values and standard errors. The log-likelihood reported by \citeA{rothe2017question} is also shown for reference.\footnote{\citeA{rothe2017question} mention that the log-likelihood values of their model are approximated and rely on an estimated partition function, which could contribute to the substantial difference compared to our model.}

The full model out-performs models without pre-training or fine-tuning by a large margin. This demonstrates that (1) the automated pre-training procedure is effective in conveying the task and DSL to the model; (2) the low log-likelihood for the ``no finetune'' model shows that the distribution of the constructed questions and real human questions are very different, which is consistent with prior work on the informativeness of human questions \cite{Rothe2018}.

The model without an encoder performs surprisingly well, but there are also reasons to think that context is not the most important factor. Some stereotyped patterns of question asking are effective across a wide range of scenarios, especially when little is known about the board (e.g., all games have the same optimal first question when no information is revealed yet). 
To further examine the role of context, we calculated the entropy of the hypothesis space of possible ship locations for each board, and group the top 5 and bottom 5 boards into high and low entropy groups. Then we calculated the average log-likelihood on different entropy groups and list the results in Table \ref{tab:exp2_result}.  When the game entropy is high, questions like ``how long is the red ship'' are good for almost any board, so the importance of the encoder is reduced. When the game entropy is low, the models with access to the board has substantially higher log-likelihood than the model without the encoder. Also, note that the first experiment would be impossible to perform well without an encoder. Together, this shows the importance of modeling the context-sensitive characteristics of how people ask questions.

\subsection{Question generation} \label{sec:exp_generation}

In this experiment, we evaluate our reinforcement learning framework on its ability to generate novel questions from scratch, without training on human-generated questions. As described before, the neural network selects a sequence of grammar-based actions to generate a question, and the model is optimized with REINFORCE \cite{williams1992simple}.

\begin{table}[h]
\centering
\footnotesize
\caption{Most frequent questions asked by humans and the grammar RL model (SP0.02) along with their frequencies.}
\begin{tabular}{llr} \toprule
\multirow{10}{*}{human} & (size Red) & 12.1\%  \\
& (size Purple) & 9.6\% \\
& (size Blue) & 9.1\% \\
& (== (orient Blue) H) & 7.0\% \\
& (== (orient Red) H) & 4.9\% \\
& (== (orient Purple) H) & 4.2\% \\
& (topleft (coloredTiles Red)) & 3.2\% \\
& (== (size Purple) 4) & 1.9\% \\
& (== (size Blue) 3) & 1.9\% \\
& (== (size Blue) 4) & 1.7\% \\ \midrule
\multirow{10}{*}{RL} & (size Red) & 19.0\% \\
& (size Blue) & 18.1\% \\
& (size Purple) & 12.3\% \\
& (orient  Red) & 8.9\% \\
& (orient Blue) & 6.9\% \\
& (orient Purple) & 4.2\% \\
& (bottomright (coloredTiles Red)) & 3.5\% \\
& (bottomright (coloredTiles Blue)) & 3.3\% \\
& (bottomright (coloredTiles Purple)) & 3.1\% \\
& (setSize (coloredTiles Red)) & 1.9\% \\ \bottomrule
\end{tabular}
\label{tab:exp3_question_dist}
\end{table}

To accomplish this, we use a reward function for training the RL agent that is based on EIG, since it is a good indicator of question informativeness and easy to compute. We give a reward of $1$ if the generated question has EIG $>0.95$, a reward of $0$ for questions with EIG $\leq0.95$, and a reward of $-1$ for invalid questions (e.g. longer than $80$ tokens). We do not directly use the EIG value as our reward because preliminary experiments show that it causes the model to choose from a very small set of high EIG questions. Instead, we wish to study how a model, like people, can generate a diverse set of good questions in simple goal-directed tasks.
To further encourage more human-like behavior, the reward function includes a step penalty for each action.

\subsubsection{Results and discussion}
We compare our program-based framework with a simple text-based model, which has the same architecture but is trained with supervision on the text-form questions.
The RL model is also compared with the program-based supervised model from the last experiment, and other RL models with different step penalties. The models are evaluated on $1000$ random boards, and generate one question for each board. The results are shown in Table \ref{tab:exp3_result}.

First, when the text-based model is evaluated on new contexts, $96.3\%$ of the questions it generates were included in the training data. We calculated the EIG of the program form of the text questions, and find that the average EIG and the ratio of EIG$>$0 is worse than the supervised model trained on programs. Some of these deficiencies are due to the very limited text-based training data, but using programs instead can help overcome these limitations. With the program-based framework, we can sample new boards and questions to create a much larger dataset with executable program representations. This self-supervised training helps to boost performance.

From Table \ref{tab:exp3_result}, the grammar-enhanced RL model is able to generate more informative and creative questions compared to the alternatives. It can be trained from scratch without examples of human questions, and produces many high EIG questions that are genuinely novel (not present in the human corpus). In contrast, the supervised model can also generate novel questions, although many have limited utility  (only $45.70$\% questions have EIG$>$0.95).
The step penalty helps the RL model to take fewer actions and to generate shorter questions. As shown in the table, model with step penalty 0.05 generates questions with on average $5.79$ tokens, but with limited diversity.
On the other hand, the RL model without a step penalty has high average EIG and the most diverse set of questions; however, it also generates many meaningless questions (only $92.40\%$ have EIG$>$0).

Another interesting finding is that questions generated by our RL agent are surprisingly consistent with human questions, even though the RL agent is not trained on any human examples. Table \ref{tab:exp3_question_dist} lists the top 10 frequent questions asked by humans and our RL model with step penalty $0.2$. Both humans and our RL model most frequently ask questions about the size of the ships, followed by questions about ship orientation. A notable difference is that people often ask true/false questions, such as ``Is this size of the Blue ship 3?'', while the RL model does not since these questions have lower EIG.

\begin{figure*}[t]
    \centering
    \includegraphics[width=\textwidth]{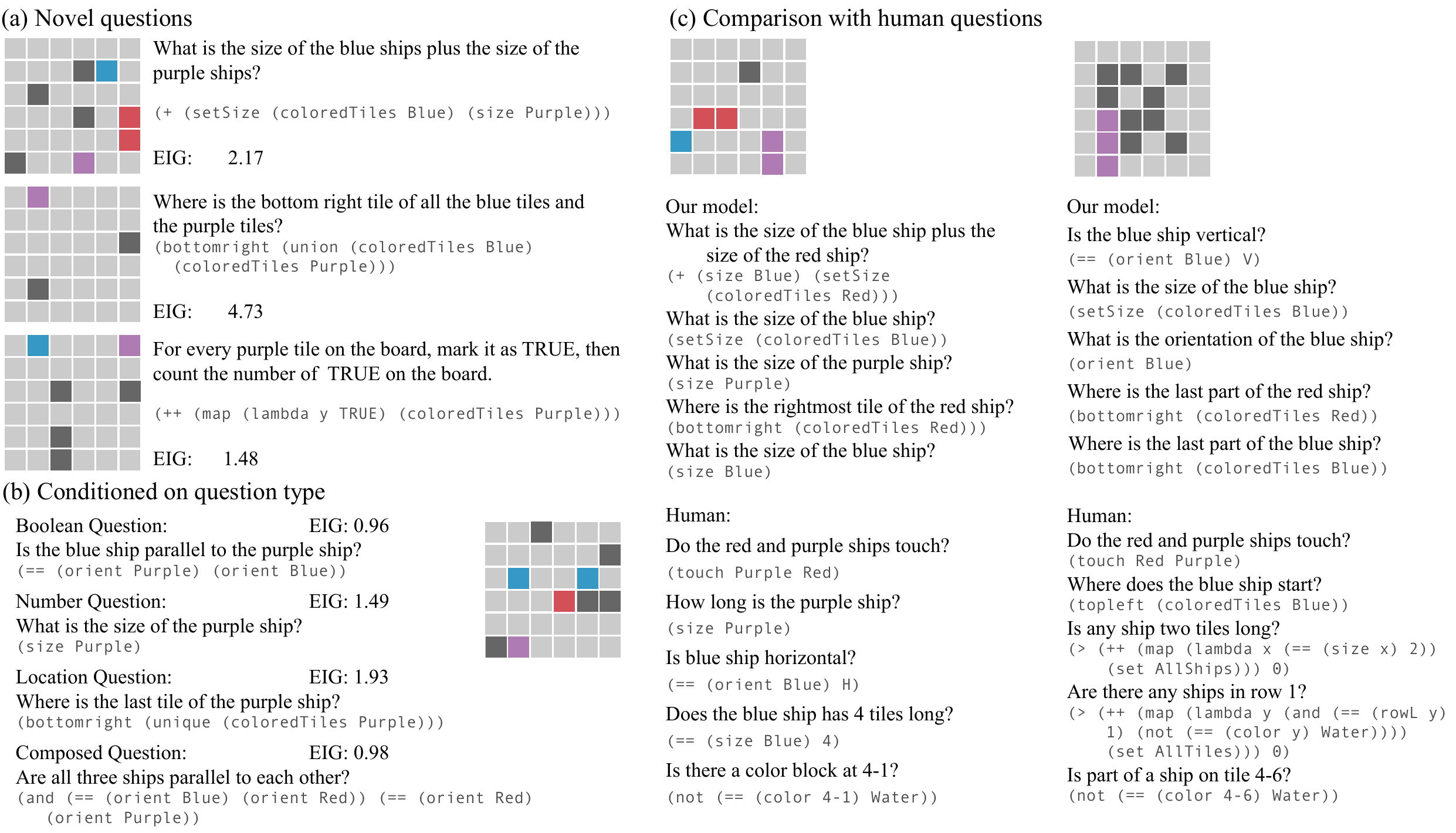}
    \caption{Examples of model-generated questions. The natural language translations of the question programs are provided for interpretation. (a) shows three novel questions generated by the RL model, (b) shows an example of the model generating different type of questions by conditioning on the decoder, (c) shows questions generated by our model as well as humans.}
    \label{fig:exp3_example}
\end{figure*}

Figure \ref{fig:exp3_example} provides examples to show the diversity of questions generated by our RL model. Figure \ref{fig:exp3_example}(a) shows novel questions produced by the model, including clever and human-like questions such as ``What is the size of the blue ship plus the purple ship?'' or ``Where is the bottom right tile of all the blue and purple tiles?'' Sometimes it also generates complex-looking questions that can actually be expressed with simpler forms, especially when using map and lambda operators. For instance, the third example in Figure \ref{fig:exp3_example}(a) is essentially equivalent to ``What is the size of the purple ship?''

With the grammar-enhanced framework, we can also guide the model to ask different types of questions, consistent with the goal-directed nature and flexibility of human question asking. The model can be queried for certain types of questions by providing different start conditions to the model. Instead of starting the derivation from the start symbol ``{\tt A}'', we can start the derivation from an intermediate state such as ``\texttt{B}'' if the model is asked for a true/false question, or a more complicated ``\texttt{(and B B)}'' if asked for a true/false question that uses ``and''.
In Figure \ref{fig:exp3_example}(b), we show examples where the model is asked to generate four specific types of questions: true/false questions, number questions, location-related questions, and compositional true/false questions. We see that the model can flexibly adapt to new constraints and generate meaningful questions that follow these constraints.

In Figure \ref{fig:exp3_example}c, we compare arbitrary samples from the model and people in the question-asking task.
These examples again suggest that our model is able to generate clever and human-like questions. There are also meaningful differences; people often ask true/false questions as mentioned before, and people sometimes ask questions with quantifiers such as ``any'' and ``all'', which are operationalized in program form with lambda functions. These questions are complicated in representation and not favored by our model.

\section{Conclusion}

We introduce a neural program generation framework for question asking in partially observable settings, which can generate creative human-like questions based on human demonstrations through supervised learning or without demonstrations through reinforcement learning. Programs provide models with a ``machine language of thought'' for compositional synthesis, and neural networks provide an efficient means of question generation. We demonstrate the effectiveness of our method through extensive experiments covering a range of human question asking abilities. 

The current model is limited in several important ways. It cannot generalize to systematically different scenarios than it was trained on, and it sometimes generates meaningless questions. We plan to further explore the model's compositional abilities in future work.
Another promising direction is to train models jointly to both ask and answer questions, which could install a richer sense of the question semantics. 

\subsubsection*{Acknowledgments}
This work was supported by Huawei. We are grateful to Todd Gureckis and Anselm Rothe for helpful comments and conversations. We thank Jimin Tan for his work on the initial version of the RL-based training procedure.

\bibliographystyle{apacite}

\setlength{\bibleftmargin}{.125in}
\setlength{\bibindent}{-\bibleftmargin}

\bibliography{refs}

\end{document}